\documentclass[sigconf]{acmart}

\AtBeginDocument{%
  \providecommand\BibTeX{{%
    \normalfont B\kern-0.5em{\scshape i\kern-0.25em b}\kern-0.8em\TeX}}}

\copyrightyear{2022}
\acmYear{2022}
\setcopyright{acmcopyright}\acmConference[SIGIR '22]{Proceedings of the 45th
International ACM SIGIR Conference on Research and Development in Information
Retrieval}{July 11--15, 2022}{Madrid, Spain}
\acmBooktitle{Proceedings of the 45th International ACM SIGIR Conference on
Research and Development in Information Retrieval (SIGIR '22), July 11--15, 2022,
Madrid, Spain}
\acmPrice{15.00}
\acmDOI{10.1145/3477495.3531992}
\acmISBN{978-1-4503-8732-3/22/07}

\usepackage{multirow}
\usepackage{makecell}
\usepackage{subfigure}
\usepackage{bm}
\usepackage{booktabs}
\usepackage{pifont}
\newcommand{\xmark}{\ding{55}}%

\newtheorem{remark}{\noindent \textbf{Remark}}

\def\mW{{\bm{W}}}
\def\mQ{{\bm{Q}}}
\def\mK{{\bm{K}}}
\def\mV{{\bm{V}}}

\newcommand{\wo}{\mW_o}

\begin{document}
\fancyhead{}

\title{Hybrid Transformer with Multi-level Fusion for \\ Multimodal Knowledge Graph Completion}







\author{Xiang Chen}
\author{Ningyu Zhang}
\authornotemark[1]
\affiliation{%
  \institution{Zhejiang University \\ AZFT Joint Lab for Knowledge Engine \\ Hangzhou Innovation Center}
  \city{Hangzhou}
  \state{Zhejiang}
  \country{China}
}
\email{xiang_chen@zju.edu.cn}
\email{zhangningyu@zju.edu.cn}

\author{Lei Li}
\author{Shumin Deng}
\affiliation{%
  \institution{Zhejiang University \\ AZFT Joint Lab for Knowledge Engine \\ Hangzhou Innovation Center}
  \city{Hangzhou}
  \state{Zhejiang}
  \country{China}
}
\email{leili21@zju.edu.cn}
\email{231sm@zju.edu.cn}

\author{Chuanqi Tan}
\affiliation{%
  \institution{Alibaba Group}
  \city{Hangzhou}
  \state{Zhejiang}
  \country{China}
}
\email{chuanqi.tcq@alibaba-inc.com}

\author{Changliang Xu}
\affiliation{%
  \institution{State Key Laboratory of Media Convergence Production Technology and Systems}
  \state{Beijing}
  \country{China}
}
\email{xu@shuwen.com}

\author{Fei Huang}
\author{Luo Si}
\affiliation{%
  \institution{Alibaba Group}
  \city{Hangzhou}
  \state{Zhejiang}
  \country{China}
}
\email{f.huang@alibaba-inc.com}
\email{luo.si@alibaba-inc.com}

\author{Huajun Chen}
\authornote{Corresponding author.}
\affiliation{%
  \institution{Zhejiang University \\ AZFT Joint Lab for Knowledge Engine \\ Hangzhou Innovation Center}
  \city{Hangzhou}
  \state{Zhejiang}
  \country{China}
}
\email{huajunsir@zju.edu.cn}

\renewcommand{\shortauthors}{Anonymous Authors}
\newcommand{\ours}{MKGformer}

\renewcommand{\shorttitle}{Multimodal Knowledge Graph Completion}

\begin{abstract}
Multimodal Knowledge Graphs (MKGs), which organize visual-text factual knowledge, have recently been successfully applied to tasks such as information retrieval, question answering, and recommendation system. Since most MKGs are far from complete, extensive knowledge graph completion studies have been proposed focusing on the multimodal entity, relation extraction and link prediction. However, different tasks and modalities require changes to the model architecture, and not all images/objects are relevant to text input, which hinders the applicability to diverse real-world scenarios. In this paper, we propose a hybrid transformer with multi-level fusion to address those issues. Specifically, we leverage a hybrid transformer architecture with unified input-output for diverse multimodal knowledge graph completion tasks. Moreover, we propose multi-level fusion, which integrates visual and text representation via coarse-grained prefix-guided interaction and fine-grained correlation-aware fusion modules. We conduct extensive experiments to validate that our {\ours} can obtain SOTA performance on four datasets of multimodal link prediction, multimodal RE, and multimodal NER\footnote{Code is available in \url{https://github.com/zjunlp/MKGformer}.}.

\end{abstract}

\ccsdesc[500]{Information systems~Information extraction}
\ccsdesc[500]{Information systems~Multimedia content creation}

\keywords{knowledge graph completion; multimodal; relation extraction; named entity recognition}



\maketitle

\section{Introduction}

\begin{figure}[!htbp]
    \centering
    \subfigure[Examples of Multimodal Link Prediction]{
        \setlength{\belowcaptionskip}{-0.5cm} 
        \includegraphics[width=0.4\textwidth]{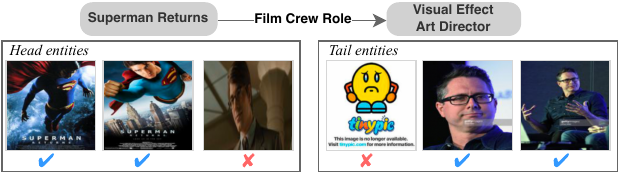} 
        \label{fig:intro1}
    }

    \subfigure[Examples of Multimodal Relation Extraction]{
        \setlength{\belowcaptionskip}{-0.5cm} 
        \includegraphics[width=0.4\textwidth]{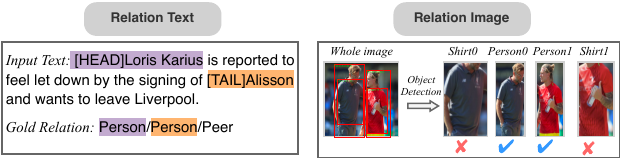}  
        \label{fig:intro2}
    }
    
    \caption{Illustration of examples of multimodal knowledge graph facts. Imaged/objects with \textcolor[RGB]{51,153,255}{\ding{52}} around the same entity have relevant visual features. In contrast, the other images/objects
    with \textcolor[RGB]{255,102,102}{\xmark} are irrelevant with the corresponding entity. }
    \label{fig:motivation}
\end{figure}

Knowledge graphs (KGs) can provide back-end support for a variety of knowledge-intensive tasks in real-world applications, such as recommender systems \cite{KSR,alicg}, information  retrieval~\cite{DBLP:conf/sigir/DietzKM18,DBLP:conf/sigir/Yang20} and time series forecasting \cite{DBLP:conf/www/DengZZCPC19}.
Since KGs usually contain visual information, Multimodal Knowledge Graphs (MKGs) recently have attracted extensive attention in the community of multimedia, natural language processing and knowledge graph~\cite{MMKG,DBLP:conf/cikm/SunCZWZZWZ20}.
However, most MKGs are far from complete due to the emerging entities and corresponding relations. 
Therefore, multimodal knowledge graph completion (MKGC), which aims to extract knowledge from the text and image to complete the missing facts in the KGs, has been proposed \cite{DBLP:conf/naacl/LiuFTCZHG21,docunet,knowprompt,DBLP:journals/corr/abs-2210-08821}. 
Concretely, visual data (images) can be regarded as complementary information used for  MKGC tasks, such as multimodal link prediction, multimodal named entity recognition (MNER) and multimodal relation extraction (MRE).

For example, as shown in Figure~\ref{fig:motivation}, 
for the multimodal link prediction task, each entity possesses many associated images, which can enhance the entity representation for missing triple prediction;
while for MNER and MRE, each short sentence contains a corresponding image to complement textual contexts for entity and relation extraction.
Benefit from the development of multimodal representation learning \cite{DBLP:journals/access/GuoWW19},
it is intuitive to fuse the heterogeneous features of KG entities and the visual information with similar semantics in unified embeddings.

To this end, \citet{IKRL} propose to integrate image features into the typical KG representation learning model for multimodal link prediction.
Besides, \citet{DBLP:conf/starsem/SergiehBGR18} and \citet{TransAE} jointly encode and fuse the visual and structural knowledge for multimodal link prediction through simple concatenation and auto-encoder, respectively.
On the other hand, 
\citet{multimodal-re} present an efficient modality alignment strategy based on scene graph for the MRE task.
\citet{zhang-UMGF} fuse regional image features and textual features with extra co-attention layers for the MNER task.

Although previous studies for multimodal KGC have shown promising improvements compared with the unimodal methods, those approaches still suffer from several evident limitations as follows:
(1) \textbf{Architecture universality}. 
Different  MKGC tasks and modality representation demand changes to the model architecture.
Specifically, different subtasks require task-specific, separately-parameterized fusion module on top of diverse encoder architectures  (e.g., ResNet~\cite{resnet}, Fast-RCNN~\cite{ren2015faster} for visual encoder, and BERT~\cite{devlin2018bert}, word2vec~\cite{DBLP:journals/corr/abs-1301-3781} for textual encoder).
Therefore, a unified model should be derived to expand the application of the diverse subtasks of multimodal KGC more effectively and conveniently.
(2) \textbf{Modality contradiction}. 
Most existing multimodal KGC models largely ignore the noise of incorporating irrelevant visual information, which may result in modality contradiction.
To be specific, in most multimodal KG, each entity possesses many associated images; however, parts of images may be irrelevant to entities, and some images even contain a lot of background noise which may mislead entity representation.
For example in Figure~\ref{fig:intro1}, each entity has many associated images, but the third image of the head entity has little relevance to the semantic meaning of  ``Superman Returns'' for multimodal link prediction.
Meanwhile, current SOTA methods for MNER and MRE tasks usually utilize valid visual objects by selecting top salient objects with the higher object classification scores, which may also introduce noise from irrelevant or redundant objects, such as the ``Shirt0'' and ``Shirt1'' objects in the Figure~\ref{fig:intro2}.
In practice, irrelevant images/objects may directly exert adverse effects on multimodal KGC.

To overcome the above barriers, we propose \textbf{\ours}, a hybrid transformer for unified multimodal KGC,
which implements the modeling of the multimodal features of the entity cross the \textbf{last few  layers} of visual transformer and the textual transformer with \textbf{multi-level fusion}, namely \textbf{M-Encoder}.
Previous works~\cite{DBLP:journals/corr/abs-2104-08696,prefix} indicate that the pre-trained models (PLMs) can activate knowledge related to the input at the self-attention layer and  Feed-Forward Network (FFN) layer in Transformer Encoder. 
Inspired by this, we consider the visual information as supplementary knowledge and propose multi-level fusion at the transformer architecture.

Specifically, we first present a coarse-grained \textbf{prefix-guided
interaction module} at the self-attention part of M-Encoder to pre-reduce modal heterogeneity for the next step.
Second, the \textbf{correlation-aware fusion module} is proposed in the FFN part of M-Encoder to obtain the fine-grained image-text representations which can alleviate the error sensitivity of irrelevant images/objects.
In particular, apart from multimodal link prediction, {\ours} can be more generally applied to  MRE and MNER tasks with a simple modification of task-specific head as shown in Figure~\ref{fig:arc1}.
In a nutshell, the contributions of this paper can be summarized:

\begin{itemize}
    \item 
    To the best of our knowledge, our work is the first to propose a hybrid transformer framework that can be applied to multiple multimodal KGC tasks.
    Intuitively, leveraging a unified transformer framework with similar arithmetic units to encode text descriptions and images inside transformers naturally reduces the heterogeneity to model better multimodel entity representation. 

    \item 
    We propose multi-level fusion with coarse-grained prefix-guided interaction module and fine-grained correlation-aware fusion module in blocks of transformers to pre-reduce the modal heterogeneity and alleviate noise of irrelevant visual elements, respectively, which are empirically necessary for diverse MKGC tasks.
   
    \item 
    We perform comprehensive experiments and extensive analysis on three tasks involving multimodal link prediction, MRE and MNER. 
    Experimental results illustrate that our model can effectively and robustly model the multimodal representations of descriptive text and images and substantially outperform the current state-of-the-art (SOTA) models in standard supervised and low-resource settings.

\end{itemize}

\section{Related Works}

\subsection{Multimodal Knowledge Graph Completion}

Multimodal KGC  has been widely studied in recent years, which leverages the associated images to represent relational knowledge better.
Previous studies mainly focus on the following three tasks: 

\subsubsection{Multimodal Link Prediction}
Existing methods for multimodal link prediction focus on encoding image features in KG embeddings.
\citet{IKRL} extend TransE~\cite{Bordes:TransE} to obtain visual representations that correspond to the KG entities and structural information of the KG separately.
\citet{DBLP:conf/starsem/SergiehBGR18},\citet{mose} and \citet{TransAE} further propose several fusion strategy to  encoder the visual and structural features into unified embedding space. 
Recently, \citet{RSME} study the noise from irrelevant images corresponding to entities and designs a forget gate with an MRP metric to select valuable images for multimodal KGC.

\subsubsection{Multimodal Relation Extraction}

Recently, \citet{9428274} present a multimodal  RE dataset with baseline models. 
The experimental results illustrate that utilizing multimodal information improves RE performance in social media texts.
\citet{multimodal-re} further revise the multimodal RE dataset and presents an efficient alignment strategy with scene graphs for textual and visual representations.
\citet{FL-MSRE} also present four multimodal datasets to handle the lack of multimodal social relation resources and propose a few-shot learning based approach to extracting social relations from both texts and face images.

\begin{figure*}[!t]
\centering
        \subfigure[Unified Multimodal KGC Framework.]{
    \includegraphics[width=0.58\textwidth]{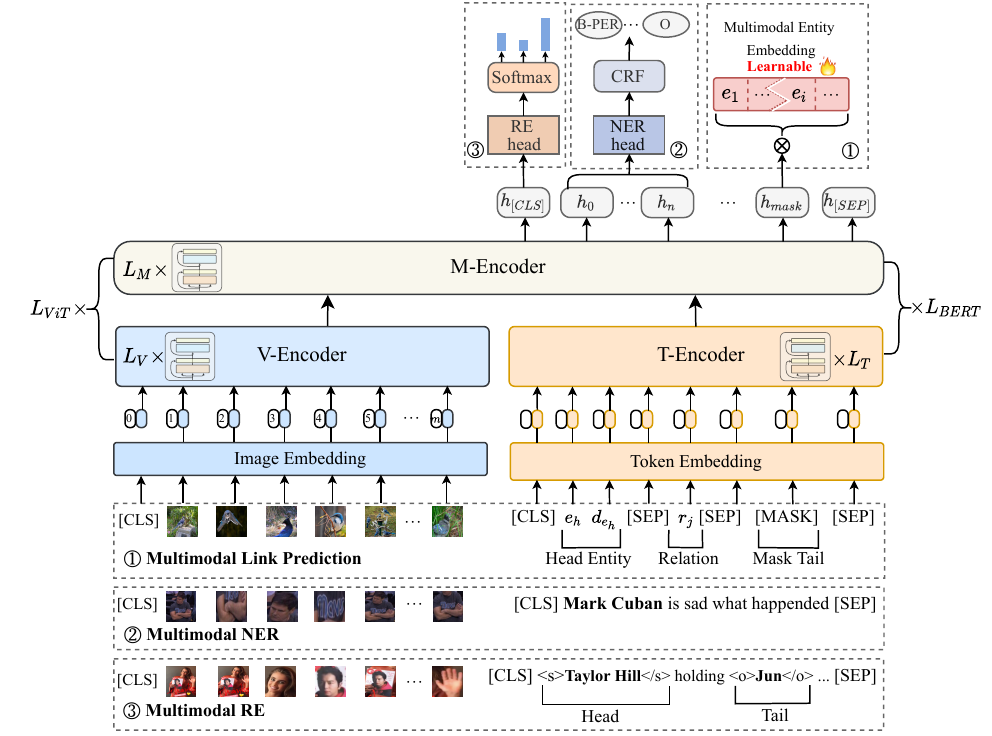}
    \label{fig:arc1}
    }
        \subfigure[Detailed M-Encoder.]{
    \includegraphics[width=0.37\textwidth]{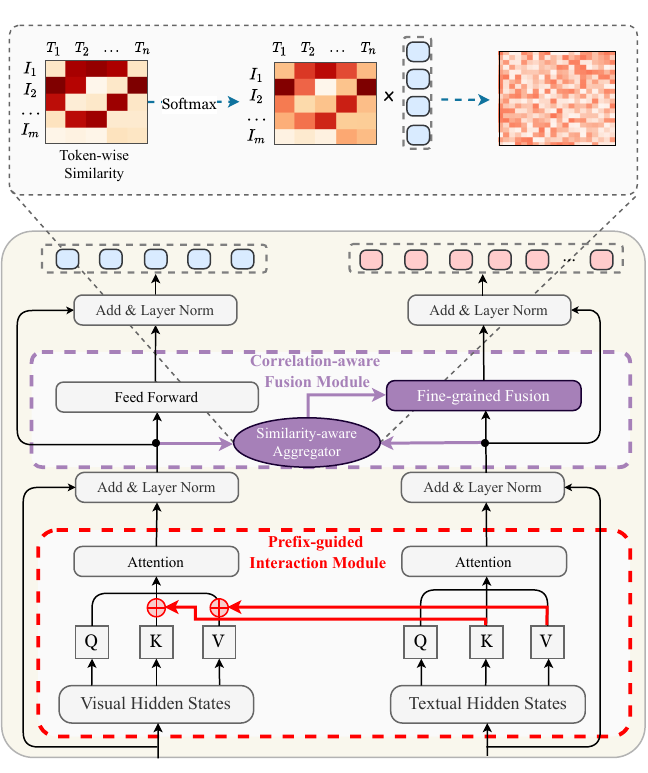}
    \label{fig:arc2}
    }
\caption{Illustration of {\ours} for (a) Unified Multimodal KGC Framework and (b) Detailed M-Encoder.}
\label{fig:arc}
\end{figure*}

\subsubsection{Multimodal Named Entity Recognition}
\citet{zhang2018adaptive}, \citet{lu2018visual}, \citet{moon2018multimodal} and ~\citet{arshad2019aiding}
 propose to encode the textual information with RNN and model the whole image representation through CNN in the early stages.
Recently, \citet{yu-etal-2020-improving,zhang-UMGF} propose to leverage regional image features to represent objects to exploit ﬁne-grained semantic correspondences based on transformer and visual backbones since informative object features are more important than the whole images for MNER tasks.
\citet{sun2021rpbert} propose a text-image relation propagation-based multimodal BERT, namely RpBERT, to reduce the interference from whole images.
However, RpBERT only focuses on the irrelevance of the whole image but ignores the noise brought by irrelevant objects.

In conclusion, MKGC can handle the problem of extending a KG with missing triples, thus, have received significant attention. 
However, different tasks and modalities demand changes to the model architecture, hindering the applicability of diverse real-world scenarios.
Therefore, we argue that a unified model should be derived to expand the application of the diverse tasks of multimodal KGC more effectively and conveniently.

\subsection{Pre-trained Multimodal Representation} 
The pre-trained multimodal visual-language models have recently demonstrated great superiority in many multimodal tasks (e.g., image-text retrieval and visual question answering ). 
The existing visual-linguistic pre-trained models can be summarized as two aspects: 
1) \textbf{Architecture}. 
The single-stream structures include VL-BERT~\cite{su2019vl}, VisualBERT~\cite{li2019visualbert}, Unicoder-VL~\cite{li2020unicoder}, and UNITER~\cite{chen2020uniter}, where the image and textual embeddings are combined into a sequence and fed into transformer to obtain contextual representations.
While two-stream structures separate visual and language processing into two streams with interacting through cross-modality or co-attentional transformer layers, which includes LXMERT~\cite{tan2019lxmert} and ViLBERT~\cite{lu2019vilbert}.
2) \textbf{Pretraining tasks}. 
The pretraining tasks of multimodal models usually involve masked language modeling (MLM), masked region classification (MRC), and image-text matching (ITM).
However, the ITM task is based on the hypothesis that the text-image pairs in the caption datasets are highly related; however, there are much noise brought by irrelevant images/objects, thus not being completely satisfied with the ITM task.
Further, most of the above models are pre-trained on the datasets of image caption, such as the Conceptual Captions~\cite{sharma-etal-2018-conceptual} or COCO caption dataset~\cite{chen2015microsoft} or visual question answering datasets.
Thus, the target optimization objects of the above pre-trained multimodal models are less relevant to multimodal KGC tasks. 

Therefore, directly applying these pre-trained multimodal methods to the multimodal KGC may not produce a good performance since multimodal KGC mainly focuses on leveraging visual information to enhance the text rather than on relying on information of both sides.
Unlike previous methods that focus on learning pre-trained multimodal representation, we regard the image as supplementary information for knowledge graph completion and propose a hybrid transformer with multi-level fusion.

\section{our approach}

In this section, we present the overall framework of {\ours}, which is a general framework that can be applied to widespread multimodal KGC tasks. 
To facilitate understanding, we introduce its detailed implementation, including the unified multimodal KGC framework in Section~\ref{sec:framework}, the hybrid transformer architecture in Section~\ref{sec:architecture} and the detailed introduction of M-Encoder in Section~\ref{sec:m-encoder}.

\subsection{Unified Multimodal KGC Framework}
\label{sec:framework}

As shown in Figure~\ref{fig:arc1}, the unified multimodal KGC framework mainly includes hybrid transformer architecture and task-specific paradigm.
Specifically, we adopt ViT and BERT as visual transformer and textual transformer models, respectively and conduct the modeling of the multimodel representations of the entity across the last  $L_M$ layers of transformers.
We introduce its detailed implementation of task-specific paradigm in the following parts.

\subsubsection{Applying to Multimodal Link Prediction}
Multimodal Link Prediction is the most popular task for multimodal KGC, which focuses on predicting the tail entity given the head entity and the query relation, denoted by $(e_h, r, ?)$.  And the answer is supposed to be always within the KG. 
In terms of the images $I_h$ that related to the entity $e_h$, we propose to model the distribution over the tail entity $e_t$ as $p (e_t | (e_h, r, I_h))$.
As shown in Figure~\ref{fig:arc1}, to fully leverage the advantage of pre-trained models, we design the specific procedure for link prediction similar to masked language modeling of pre-trained language models (PLMs).
We take first step to model the image-text incorporated entity representations and then predict the missing entity $(e_h, r, ?)$ over the multimodal entity representations.

\paragraph{\textbf{Image-text Incorporated Entity Modeling}}
Unlike previous work simply concatenate or fuse based on particular visual and textual features of entities, we fully leverage the  "masked language modeling" (MLM) ability of pre-trained transformers to model image-text incorporated multimodal representations of the entities in the knowledge graph.
To be more specific, given an entity description $d_{e_i} = (w_1,...,w_n)$ and its corresponding multiple images $I_{e_i}=\{I^1, I^2, ..., I^o\}$, 
we feed the patched images of the entity $e_i$ into the visual side of  hybrid transformer architecture and  convert the textual side input sequence of hybrid transformer architecture  as:
\begin{equation}
\label{eq:entity_modeling}
T_{e_i}=\texttt{[CLS]} d_{e_i} \ \texttt{is} \ \texttt{the} \ \texttt{description} \ \texttt{of} \ \texttt{[MASK]}  \texttt{[SEP]}.
\end{equation} 
We extend the word embedding layer of BERT to treat each token embedding as corresponding multimodal representation $E_{e_i}$ of $i$-th entity $e_i$.
Then we train the {\ours} to predict the $\texttt{[MASK]}$  over the multimodal entity embedding $E_{e_i}$ with cross entropy loss for classification:
\begin{equation}
\label{eq:entity_predict}
\small
\mathcal L_{link} =- log(p(e_i|(T_{e_i})),
\end{equation} 
Notably, we freeze the whole model except the newly added parameters of multimodal entity embedding.
We argue that the modified input can guide {\ours} to incorporate the textual and visual information into multimodal entity embeddings attentively.

\paragraph{\textbf{Missing Entity Prediction}}
Given a $triple = (e_h, r,e_t) \in \mathcal{G}$, KGC models predict the $e_h$ or $e_t$ in the head or tail batch.
Similarly, we treat the link prediction as the MLM task, which uses entity $e_h$, entity description $d_{e_h}$, relation $r$ and the entity images $I_{e_h}$ to predict the masked tail entity over the multimodal entity embeddings described above.
Specifically, we also process the multiple patched images of the entity $e_h$ into the visual side of  hybrid transformer architecture and  convert this triple $(e_h, r, ?)$ to the input sequence of on the text side as follows:
\begin{equation}
\label{eq:lp_predict}
T_{(e_h, r, ?)}=\texttt{[CLS]}  e_h \ d_{e_h} \texttt{[SEP]} r_j \texttt{[SEP]} \texttt{[MASK]} \texttt{[SEP]}.
\end{equation} 
Finally, we train the {\ours} to predict the $\texttt{[MASK]}$  over the multimodal entity embedding $E_{e_i}$ via binary cross-entropy loss for multilabel classification with the consideration that the prediction of $e_t$ is not unique in link prediction.

\subsubsection{Applying to MRE}
Relation extraction aims at linking relation mentions from text to a canonical relation type in a knowledge graph. 
Given the text $T$ and the corresponding image $I$, we aim to predict the relation between an entity pair $(e_h, e_t)$ and outputs the distribution over relation types as  $p (r |  (T,I, e_h, e_t))$.
Specifically, we take the  representation of the special token $\texttt{[CLS]}$ from the final output embedding of hybrid transformer architecture to compute the probability distribution over the class set $\mathcal{Y}$ with the softmax function $p (r | ( I, T, e_h, e_t)) = \texttt{Softmax}(\mathbf{W} {\bm{h}_{L_M}^{M_t}}_{\texttt{[CLS]}})$. 
${\bm{h}_{L_M}^{M_t}} \in{\mathbb{R}^{n\times{d_T}}}$ denotes the final sequence representation of the $L_M$-th layer from textual side of M-Encoder in hybrid transformer architecture.
The parameters of the model and $\mathbf{W}$ are fine-tuned by minimizing the cross-entropy loss over $p (r | ( I, T, e_h, e_t))$ on the entire train sets.

\subsubsection{Applying to MNER}
MNER is the task of extracting  named entities from text sequences and corresponding images. 
Given a token sequence  $T=\{w_1, \ldots, w_{n}\}$ and its corresponding image $I$, we focus on modeling the distribution over sequence tags as $p (y | (T, I))$, where $y$ is the label sequence of tags $y= \{y_1,\ldots,y_n\}$.
We assign the procedure of {\ours} with CRF~\cite{lample2016neural} function similar to previous multimodal NER tasks for the fair comparison. For a sequence of tags $y= \{y_1,\ldots,y_n\}$, we calculate the probability of the label sequence $y$ over the pre-deﬁned label set $Y$  with the BIO tagging schema as described  in~\cite{lample2016neural}.

\subsection{Hybrid Transformer Architecture}
\label{sec:architecture}
The \textbf{hybrid transformer architecture} of {\ours} mainly includes  three stacked modules: (1) the underlying textual encoder is designed to capture basic syntactic and lexical information from the input tokens, namely ($\textbf{T-Encoder}$),
(2) the underlying visual encoder ($\textbf{V-Encoder}$), which is  responsible for capturing basic visual features from the input patched images,
and (3) the upper multimodal encoder ($\textbf{M-Encoder}$) is adopted to model image-text incorporated entity representations inside the underlying visual transformer and textual transformer.
Besides, we denote the number of $\texttt{V-Encoder}$ layers as $L_V$, the number of $\texttt{T-Encoder}$ layers as $L_T$, and the number of $\texttt{M-Encoder}$ layers as $L_M$, where $L_{ViT}=L_V+L_M$ and $L_{BERT}=L_T+L_M$.

\paragraph{\textbf{Recap of the Transformer Architecture}}
\label{sec:transformer}
Transformer ~\citep{vaswani2017attention} is now the workhorse architecture behind most SOTA models in CV and NLP, which is composed of $L$ stacked blocks.
While each block mainly includes two types of sub-layers: multi-head self-attention (MHA) and a fully connected feed-forward network (FFN).
Layer Normalization (LN) and residual connection are also used in each layer.
Given the input sequence vectors $\bm{x}\in\mathbb{R}^{n \times d} $,
the conventional attention function maps $\bm{x}$ to queries $\mQ \in \mathbb{R}^{n \times d}$ 
and key-value pairs $\mK \in \mathbb{R}^{n\times d}, \mV \in \mathbb{R}^{n\times d}$:

\begin{equation}
    \mathrm{Attn}(\mQ, \mK, \mV) = \text{softmax}(\frac{\mQ\mK^T}{\sqrt{d}})\mV,
\end{equation}
where $n$ denotes the length of sequence.
MHA performs the attention function in parallel over $N_h$ heads, where each head is separately parameterized by $\mW_q^{(i)}, \mW_k^{(i)}, \mW_v^{(i)} \in \mathbb{R}^{d\times d_h}$ to project inputs to queries, keys, and values. 
The role of MHA is to compute the weighted hidden states for each head, and then concatenates them as:
\begin{equation}
\begin{aligned}
    \mathrm{MHA}(\bm{x}) &=  \mathrm{[head_1; \cdots; head_h ]}\wo, \\
    \mQ^{(i)}, \mK^{(i)}, \mV^{(i)} &= \bm{x} \mW_q^{(i)}, \bm{x}\mW_k^{(i)}, \bm{x}\mW_v^{(i)} \\
    \mathrm{head_i} &= \mathrm{Attn}(\mQ^{(i)}, \mK^{(i)}, \mV^{(i)}), 
    \label{eq:multihead:attn}
\end{aligned}
\end{equation}
where $\wo \in \mathbf{R}^{d\times d}$ and
$d$ denotes the dimension of hidden embeddings.  
 $d_h=d / N_h$ is typically set in MHA.
 FFN is another vital component in transformer, typically consisting of two layers of linear transformations with a ReLU activation function as follows:
\begin{equation}
    \mathrm{FFN}(\textnormal{x}) = \mathrm{ReLU}(\bm{x}\mW_1 +\mathbf{b}_1)\mW_2 + \mathbf{b}_2,
\end{equation}
where $\mW_1 \in \mathbb{R}^{d\times d_m}$, $\mW_2 \in \mathbb{R}^{d_m\times d}$.  

\paragraph{\textbf{V-Encoder.}}
We adopt the first $L_V$ layers of  ViT~\cite{vit} pre-trained on ImageNet-1k from \cite{deit} as the visual encoder to extract image features.
Given $o$ images $I_{e_i}$ of the entity $e_i$\footnote{Here, we take the multimodal link prediction as example. As for multimodal NER and RE, we choose the top $o$ salient objects according to the text.}, 
we rescale each  image to unified $H \times W$ pixels, and the $i$-th input image $\textit{I} _{i}\in{\mathbb{R}^{C\times{H}\times{W}}} (1\le i< o)$  is first reshaped into $u=HW/P^2$ flattened 2D patches, then pooled and projected as $X_{pc}^{i}\in{\mathbb{R}^{u\times{d_V}}}$, where the resolution of the input image is $H\times{W}$, $C$ is the number of channels and $d_V$ denotes the dimension of hidden states of ViT. 
We concatenate the 
patched embeddings of $o$ images to get the visual sequence patch embeddings $X_{pc}\in{\mathbb{R}^{ m \times{d_V}}}$, where $m=(u\times{o})$. 

\begin{equation}
    \begin{aligned}
    X_{0}^V &= X_{pc} + V_{pos} \\ 
    \Bar{X}_{l}^{V} &= \mathbf{MHA}(\mathbf{LN}(X_{0}^V)) + X_{l-1}^{V}, l=1...L_V \\
    X_{l}^{V} &= \mathbf{FFN}(\mathbf{LN}(\Bar{X}_{l}^{V})) + \Bar{X}_{l}^{V}, l=1...L_V,
    \end{aligned}
\end{equation}
where $V_{pos}\in{\mathbb{R}^{m\times{d_V}}}$ represents the corresponding position embedding layer,  embedding, $X_{l}^{V}$ is the hidden states of the $l$ layer of visual encoder.

\paragraph{\textbf{T-Encoder.}}
We leverage the first $L_T$ layers of BERT~\cite{devlin2018bert} as the text encoder, which also consists of $L_T$ layers of MHA and FFN blocks similar to the visual encoder except that LN comes after MHA and FFN. 
To be specific, a token sequence $\{w_1, \ldots, w_n\}$ is embedded to $X_{wd}\in{\mathbb{R}^{n\times{d_T}}}$ with a word embedding matrix, and the textual representation is calculated as follows:
\begin{equation}
    \begin{aligned}
    X_{0}^T&= X_{wd}  + T_{pos} \\ 
    \Bar{X}_{l}^{T} &= \mathbf{LN}(\mathbf{MHA}(X_{l-1}^{T})) + X_{l-1}^{T}, l=1...L_T \\
    X_{l}^T &= \mathbf{LN}(\mathbf{FFN}(\Bar{X}_{l}^{T})) + \Bar{X}_{l}^{T}, l=1...L_T,
    \end{aligned}
\end{equation}
where $T_{pos}\in{\mathbb{R}^{(n)\times{d_T}}}$ denotes position embedding, $X_{l}^T $ is the hidden states of the $l$ layer for the output textual sequence.

\paragraph{\textbf{M-Encoder.}}

Multimodal KGC mainly faces the issues of heterogeneity and irrelevance between different modalities. 
Different from previous works leveraging extra co-attention layers to integrate modality information,
we propose to model the multimodal features of the entity cross the last $L_M$ layers of ViT and BERT with multi-level fusion, namely M-Encoder.
To be specific, we present a \textbf{P}refix-\textbf{G}uided \textbf{I}nteraction module (\textbf{PGI}) at the self-attention block to pre-reduce the modality heterogeneity.
We also propose a \textbf{C}orrelation-\textbf{A}ware \textbf{F}usion module (\textbf{CAF}) in the FFN layer to reduce the impact of noise caused by irrelevant image elements.
Here, we omit the calculation of LN and residual connection for simplicity. 
\begin{equation}
    \begin{aligned}
   h_{0}^{M_t} &= X_{L_T}^T& \\ 
   h_{0}^{M_v} &= X_{L_V}^V& \\ 
   \Bar{h}_{l}^{M_t}, \Bar{h}_{l}^{M_v} &=
   \mathbf{PGI}(h_{l-1}^{M_t},h_{l-1}^{M_v}), l=1...L_M \\
   h_{l}^{M_t}, h_{l}^{M_v} &=
   \mathbf{CAF}(\Bar{h}_{l-1}^{M_t},\Bar{h}_{l-1}^{M_v}), l=1...L_M .\\
    \end{aligned}
\end{equation}

\subsection{Insights of M-Encoder}
\label{sec:m-encoder}
\subsubsection{Prefix-guided Interaction Module}
\label{sec:HPM}
Inspired by the success of textual prefix tuning~\cite{prefix} and corresponding analysis~\cite{DBLP:journals/corr/abs-2110-04366}, we propose a prefix-guided interaction mechanism to pre-reduce the modality heterogeneity through the calculation of multi-head attention at every layer, which is performed on the hybrid keys and values. In particular, 
we redefine the computation of visual  $\mathrm{head}_i^{M_v}$ and textual  $\mathrm{head}_i^{M_t}$  in Eq.~\ref{eq:multihead:attn} as:
\begin{equation}
\label{eq:pt}
\small
\begin{aligned}
        \mathrm{head}^{M_t} &= \mathrm{Attn}(\bm{x}^{t}\mW_q^{t}, \bm{x}^{t}\mW_k^{t}, \bm{x}^{t}\mW_v^{t}),  \\
         \mathrm{head}^{M_v} &= \mathrm{Attn}(\bm{x}^{v}\mW_q^{v}, [\bm{x}^{v}\mW_k^{v}, \bm{x}^{t}\mW_k^{t}], 
         [\bm{x}^{v}\mW_v^{v}, \bm{x}^{t}\mW_v^{t}]),
\end{aligned}
\end{equation}
We also derive the variant formula of Eq.~\ref{eq:pt} and provide another perspective of prefix-guided interpolated attention:\footnote{Without loss of generalization, we ignore the softmax scaling factor $\sqrt{d}$ for ease of representation.}
\begin{equation}
\label{eq:prefix-adapter}
\small
\begin{split}
& \mathrm{head}^{M_v} = \mathrm{Attn}(\bm{x}^{v}\mW_q^{v}, [\bm{x}^{v}\mW_k^{v}, \bm{x}^{t}\mW_k^{t}], 
[\bm{x}^{v}\mW_v^{v}, \bm{x}^{t}\mW_v^{t}]), \\
& = \text{softmax}\big( \mQ_v [\mK_v; \mK_t] ^\top\big) \begin{bmatrix} \mV_v\\ \mV_t \end{bmatrix} \\
& = (1 - \lambda(\bm{x}^{v})) \text{softmax}(\mQ_v {\mK_v}^\top)\mV_v 
 + \lambda(\bm{x}^{v})\text{softmax}(\mQ_v {\mK_t}^\top) \mV_t \\
& = (1 - \lambda(\bm{x}^{v})) \underbrace{ \text{Attn}(\mQ_v, \mK_v, \mV_v) }_{\text{standard attention}} + \lambda(\bm{x}^{v}) \underbrace{ \text{Attn}(\mQ_v, \mK_t, \mV_t) }_{\text{Cross-modal Interaction }},
\end{split}
\end{equation}
\begin{equation}
\lambda(\bm{x}^{v}) = \frac{\sum_i\exp (\mQ_v \mK_t^\top)_i}{\sum_i \exp (\mQ_v \mK_t^\top)_i + \sum_j \exp(\mQ_v \mK_v^\top)_j}.
\end{equation} 
where $\lambda(\bm{x}^{v})$ denotes the scalar for the sum of normalized attention weights on the textual key and value vectors.

\begin{remark}
As shown in Eq.~\ref{eq:prefix-adapter}, 
the first term $\text{Attn}(\mQ_v, \mK_v, \mV_v)$ is the standard attention in the visual side, whereas the second term represent the cross-modal interaction.
Monolithic in the sense that the prefix-guided interaction mechanism down-weights the original visual attention probabilities by a scalar factor (i.e., $1-\lambda$) and redistributes the remaining attention probability mass $\lambda$ to attend to textual attention, which likes the linear interpolation.
By applying this to the attention flow calculation over hidden visual states and hidden textual states,
{\ours} learns coarse-grained modality fusion to pre-reduce the modality heterogeneity.
\end{remark}

\begin{table*}[htbp!]
  \centering
  \caption{Results of the link prediction on FB15K-237-IMG and WN18-IMG. 
  Note that the universal pre-trained vision-language model cannot be directly applied to the multimodal link prediction; thus, we follow KG-BERT to leverage the pre-trained VL model for multimodal link prediction.  
  }
    \scalebox{0.85}{
    \begin{tabular}{c|rrrr|rrrr}
   \toprule
        \multirow{2}{*}{Model}      
          & \multicolumn{4}{c|}{\textbf{FB15k-237-IMG}}
          & \multicolumn{4}{c}{\textbf{WN18-IMG}} \\
    \cmidrule{2-9} 
          & \multicolumn{1}{c}{Hits@1 $\uparrow$} 
          & \multicolumn{1}{c}{Hits@3 $\uparrow$} 
          & \multicolumn{1}{c}{Hits@10 $\uparrow$} 
          & \multicolumn{1}{c|}{MR $\downarrow$}   
         & \multicolumn{1}{c}{Hits@1 $\uparrow$} 
          & \multicolumn{1}{c}{Hits@3 $\uparrow$} 
          & \multicolumn{1}{c}{Hits@10 $\uparrow$} 
          & \multicolumn{1}{c}{MR $\downarrow$}       \\
    \midrule
    \multicolumn{9}{c}{\textit{Unimodal  approach}} \\
    \midrule
    \multicolumn{1}{c|}{TransE~\cite{Bordes:TransE}} 
    & \multicolumn{1}{c}{0.198} & \multicolumn{1}{c}{0.376} & \multicolumn{1}{c}{0.441} & \multicolumn{1}{c|}{323} 
    
   &\multicolumn{1}{c}{0.040} 
   &\multicolumn{1}{c}{0.745} &\multicolumn{1}{c}{0.923} &\multicolumn{1}{c}{357} \\
    
    \multicolumn{1}{c|}{DistMult~\cite{Yang:DistMult}}
      & \multicolumn{1}{c}{0.199} & \multicolumn{1}{c}{0.301} & \multicolumn{1}{c}{0.446} & \multicolumn{1}{c|}{512}
      
    &\multicolumn{1}{c}{0.335} 
   &\multicolumn{1}{c}{0.876} &\multicolumn{1}{c}{0.940} &\multicolumn{1}{c}{655} \\
   
    \multicolumn{1}{c|}{ComplEx~\cite{Trouillon:ComplEx}}
   & \multicolumn{1}{c}{0.194} & \multicolumn{1}{c}{0.297} & \multicolumn{1}{c}{0.450} & \multicolumn{1}{c|}{546}
 
& \multicolumn{1}{c}{0.936} & \multicolumn{1}{c}{0.945} & \multicolumn{1}{c}{0.947} & \multicolumn{1}{c}{-}
 \\
    
    \multicolumn{1}{c|}{RotatE~\cite{RotatE}} 
    & \multicolumn{1}{c}{0.241} & \multicolumn{1}{c}{0.375} & \multicolumn{1}{c}{0.533} & \multicolumn{1}{c|}{177}
    
   &\multicolumn{1}{c}{0.942} 
   &\multicolumn{1}{c}{0.950} &\multicolumn{1}{c}{0.957} &\multicolumn{1}{c}{254} \\

     \multicolumn{1}{c|}{KG-BERT~\cite{DBLP:journals/corr/abs-1909-03193}  }
     & \multicolumn{1}{c}{-} 
     & \multicolumn{1}{c}{-} 
     & \multicolumn{1}{c}{0.420} 
     & \multicolumn{1}{c|}{153}  
     
     &\multicolumn{1}{c}{0.117} 
   &\multicolumn{1}{c}{0.689} &\multicolumn{1}{c}{0.926} &\multicolumn{1}{c}{58} \\

   \midrule
    
    \multicolumn{9}{c}{\textit{Multimodal approach}} \\
    \midrule
     
  \multicolumn{1}{c|}{VisualBERT\_{base} ~\cite{li2019visualbert}} 
     &\multicolumn{1}{c}{0.217} 
  &\multicolumn{1}{c}{0.324} &\multicolumn{1}{c}{0.439} &\multicolumn{1}{c|}{592}
   
    &\multicolumn{1}{c}{0.179} 
  &\multicolumn{1}{c}{0.437} &\multicolumn{1}{c}{0.654} &\multicolumn{1}{c}{122} \\
  
   \multicolumn{1}{c|}{ViLBERT\_{base} ~\cite{vilbert}} 
   &\multicolumn{1}{c}{0.233} 
  &\multicolumn{1}{c}{0.335} &\multicolumn{1}{c}{0.457} &\multicolumn{1}{c|}{483}
   
    &\multicolumn{1}{c}{0.223} 
  &\multicolumn{1}{c}{0.552} &\multicolumn{1}{c}{0.761} &\multicolumn{1}{c}{131} \\
  
    \multicolumn{1}{c|}{IKRL(UNION)~\cite{IKRL}} 
     &\multicolumn{1}{c}{0.194} 
  &\multicolumn{1}{c}{0.284} &\multicolumn{1}{c}{0.458} &\multicolumn{1}{c|}{298}
   
    &\multicolumn{1}{c}{0.127} 
  &\multicolumn{1}{c}{0.796} &\multicolumn{1}{c}{0.928} &\multicolumn{1}{c}{596} \\

  \multicolumn{1}{c|}{TransAE~\cite{TransAE}} 
  &\multicolumn{1}{c}{0.199} 
  &\multicolumn{1}{c}{0.317} &\multicolumn{1}{c}{0.463} &\multicolumn{1}{c|}{431}
   
  &\multicolumn{1}{c}{0.323} 
  &\multicolumn{1}{c}{0.835} &\multicolumn{1}{c}{0.934} &\multicolumn{1}{c}{352} \\

    \multicolumn{1}{c|}{RSME (ViT-B/32+Forget)~\cite{RSME}} 
     &\multicolumn{1}{c}{0.242} 
  &\multicolumn{1}{c}{0.344} &\multicolumn{1}{c}{0.467} &\multicolumn{1}{c|}{417}
   
    &\multicolumn{1}{c}{0.943} 
  &\multicolumn{1}{c}{0.951} &\multicolumn{1}{c}{0.957} &\multicolumn{1}{c}{223} \\
     \midrule
    \multicolumn{1}{c|}{{\ours}} 
      & \multicolumn{1}{c}{0.256} &\multicolumn{1}{c}{0.367} 
    & \multicolumn{1}{c}{0.504}
    & \multicolumn{1}{c|}{221} 
    
    &\multicolumn{1}{c}{0.944} 
   &\multicolumn{1}{c}{0.961} &\multicolumn{1}{c}{0.972} &\multicolumn{1}{c}{28}  \\
    \bottomrule
    \end{tabular}%
    }
  \label{tab:full_lp}%
\end{table*}%

\begin{table}[!t]
    \caption{\textbf{Performance of low-resource setting  (8-shot) FB15K-237-IMG for multimodal link prediction. }}
    \label{tab:low_resource_lp}
    \centering
        \scalebox{0.85}{
    \begin{tabular}{c|cc}
    \toprule
    \multirow{2}{*}{Model}               
    & \multicolumn{2}{c}{\textbf{FB15K-237-IMG}}       \\ 
        \cmidrule{2-3} 
    & MR   $\downarrow$           & Hit@10 $\uparrow$   \\ 
    \midrule
     \multicolumn{1}{c|}{TransE~\cite{Bordes:TransE}}  & 5847   & 0.0925       \\
     \multicolumn{1}{c|}{DistMult~\cite{Yang:DistMult}}   & 5791   & 0.0059   \\
     \multicolumn{1}{c|}{ComplEx~\cite{Trouillon:ComplEx}}    & 6451   & 0.0046     \\
     \multicolumn{1}{c|}{RotatE~\cite{RotatE}}   & 7365   & 0.0066      \\
   \multicolumn{1}{c|}{KG-BERT~\cite{DBLP:journals/corr/abs-1909-03193}  }   & 2023   & 0.0451    \\
   \midrule
    \multicolumn{1}{c|}{VisualBERT\_{base} ~\cite{li2019visualbert}} & 5983   & 0.0772 \\
   \multicolumn{1}{c|}{ViLBERT\_{base} ~\cite{vilbert}} 
   & 5754   & 0.0831    \\
   
    IKRL~\cite{IKRL}   & 4913   & 0.0973    \\
    RSME(ViT-B/16+Forget)~\cite{RSME}  & {2895}   & 0.1240 \\
   
    {\ours}    & 2785   & 0.1344     \\ \bottomrule
    \end{tabular}
    }
\end{table}

\subsubsection{Correlation-aware Fusion Module} 

To alleviate the adverse effects of noise, we apply a correlation-aware fusion module to conduct the token-wise cross-modal interaction (e.g., word-patch alignment) between the two modalities.
Specifically, we denote $m$ and $n$ as the  sequence length of the visual vectors $\bm{x}^v \in \mathbb{R}^{m \times d}$  and textual vectors $\bm{x}^t \in \mathbb{R}^{n \times d}$    respectively, which are the corresponding output features of the prefix-guided interaction  module.
For the textual tokens, we compute its similarity matrix of  all visual tokens as follows:
\begin{equation}
\label{eq:late_sim_0}
   \bm{S} = \bm{x}^t {(\bm{x}^v)}^\top . 
\end{equation}
We then conduct softmax function over similarity matrix $\bm{S}$ of $i$-th textual token and use the average token-wise aggregator over visual tokens in the image as follows: 
\begin{equation}
 \mathrm{Agg}_{i}(\bm{x}^{v}) = \text{softmax}(\bm{S}_i)\bm{x}^{v},  {(1\le i < n)} \label{eq:late_sim_1}
\end{equation}
\begin{equation}
 \mathrm{Agg}(\bm{x}^{v}) = [\mathrm{Agg}_{1}(\bm{x}^{v});...,\mathrm{Agg}_{n}(\bm{x}^{v})] \label{eq:late_sim_2}
\end{equation}
where $\mathrm{Agg}_{i}$ denotes the similarity-aware aggregated visual representation for $i$-th textual token. 
Inspired by the finding \cite{DBLP:conf/emnlp/GevaSBL21} that the FFN layer learns task-specific textual patterns, we propose to incorporate similarity-aware aggregated visual hidden states into textual hidden states in FFN layers and modify the calculation of the FFN process as:
\begin{equation}
    \mathrm{FFN}(\bm{x}^t) = \mathrm{ReLU}(\bm{x}^t\mW_1 +\mathbf{b}_1 + \mathrm{Agg}(\bm{x}^{v}) \mW_3 )\mW_2 + \mathbf{b}_2,
\end{equation}
where $\mW_3 \in \mathbb{R}^{d\times d_m}$ represent the new added parameters for  aggregated visual hidden states.

\begin{remark}
Note that the token-wise similarity in Equation~\ref{eq:late_sim_0},~\ref{eq:late_sim_1} and \ref{eq:late_sim_2} indicates that we would like to obtain the closest image patch for each textual token.
By inserting the similarity-aware aggregated visual representation into the FFN calculation in the textual side,
our {\ours} learns fine-grained alignment between image patches and textual tokens, which makes our modal more robust to the noise of irrelevant images of entities in KG.
\end{remark}

\begin{table}[!t]
    \caption{Dataset statistics for Multimodal Link Prediction.}
        \label{table:lp-datasets}
    \centering
    \resizebox{0.4\textwidth}{!}{
        \begin{tabular}{cccccc}
            \toprule
            \textbf{Dataset}  & \textbf{\#Rel.} & \textbf{\#Ent.} & \textbf{\#Train}  & \textbf{\#Dev} & \textbf{\#Test}\\
            \midrule
            FB15k-237-IMG      & 237        & 14,541           & 272,115 
            & 17,535          &20,466
            \\    
            WN18-IMG      & 18        & 40,943           & 141,442
            & 5,000         &5,000
            \\    
            \bottomrule
        \end{tabular}
    }
\end{table}%

\section{Experiments}
 We next introduce the experimental settings  of {\ours} in three tasks: multimodal link prediction, multimodal RE, and multimodal NER. 
Following results show that {\ours} can outperforms the other baselines in both standard supervised and few-shot settings.

\begin{table*}[htbp]
\centering
\small
\caption{\label{tab:full_re}
Performance comparison of the different competitive baseline approaches for multimodal RE and NER. ``(CRF)'' represents CRF is only for MNER dataset Twitter-2017.
}
\scalebox{0.85}{
\begin{tabular}{c|l|ccc|ccc}
\toprule

{\multirow{2}{*}{Modality}} 
& {\multirow{2}{*}{Methods}} 
& \multicolumn{3}{c}{\textbf{MNRE}}
& \multicolumn{3}{c}{\textbf{Twitter-2017}}

\\

\cmidrule{3-8}

 &  & Precision  & Recall  & F1  & Precision  & Recall  & F1 \\

\midrule

    \multirow{5}{*}{Text} 
    & CNN-BiLSTM-(CRF)~\cite{acl-MaH16}    & 60.18  & 46.32 & 52.35 & 80.00 & 78.76 & 79.37 
    \\
    & HBiLSTM-(CRF)~\cite{naacl-LampleBSKD16}    & 60.22 & 47.13 & 52.87  & 82.69 & 78.16 & 80.37
    \\
    & PCNN~\cite{PCNN} & 62.85 & 49.69 & 55.49 & 83.28 & 78.30 & 80.72 
    \\
    & BERT-(CRF)    & 63.85 & 55.79 & 59.55
    & 83.32 & 83.57 & 83.44
    \\

    & MTB~\cite{MTB}  & 64.46 & 57.81 & 60.86   & 83.88 & 83.22 & 83.55
    \\

\midrule
    \multirow{8}{*}{Text+Image} 
    & AdapCoAtt-BERT-(CRF)~\cite{zhang2018adaptive}   & 64.67 & 57.98 & 61.14 & 85.13 & 83.20 & 84.10
    \\
    & UMT~\cite{yu-etal-2020-improving}   & 62.83 & 61.32 & 62.56 & 85.28 & 85.34 & 85.31
    \\
  
    & BERT\_{base}+SG~\cite{multimodal-re}  
     & 62.95 & 62.65 & 62.80  & 84.13 & 83.88 & 84.00
    \\

     & VisualBERT\_{base}~\cite{li2019visualbert}    & 56.34 & 58.28 & 57.29  & 84.06 & 85.39 & 84.72
     \\

    & ViLBERT\_{base}~\cite{vilbert} & 64.50 & 61.86 & 63.16 & 84.62 & 85.47 & 85.04
     \\
       & UMGF~\cite{zhang-UMGF}   & 64.38 & 66.23 & 65.29 & 86.54 & 84.50 & 85.51 \\
    & MEGA~\cite{multimodal-re} & 64.51 & 68.44 & 66.41  & 84.03 & 84.75 & 84.39
    \\
\cmidrule{2-8}

    & {\ours} 
     & 82.67  & 81.25 & 81.95 & 86.98 & 88.01 & 87.49
    \\
\bottomrule
\end{tabular}
}
\end{table*}

\subsection{Experimental Setup}
\subsubsection{{Datasets}}
We adopt two publicly available datasets for multimodal link prediction, including:
1) \textbf{WN18-IMG:} WN18~\cite{Bordes:TransE} is a knowledge graph  originally extracted from WordNet~\cite{ Miller:WordNet}.
While WN18-IMG is an extended dataset of WN18~\cite{Bordes:TransE} with 10 images for each entity.
\textbf{FB15K-237-IMG}: FB15K-237-IMG\footnote{Since the dataset FB15k has the inverse relation.
Therefore, we adopt corresponding sub-datasets FB15k-237 to mitigate the problem of the reversible relation between. The multimodal datasets of link prediction can be acquired from \url{https://github.com/wangmengsd/RSME}, which is the public code of RSME.} ~\cite{Bordes:TransE} has 10 images for each entity and is a subset of the large-scale knowledge graph Freebase~\cite{ BGF:Freebase}, which is a popular dataset in knowledge graph completion. 
Detailed statistics are shown in Table~\ref{table:lp-datasets}.
For multimodal RE, we evaluate on \textbf{MNRE}~\cite{multimodal-re}, a manually-labeled dataset for multimodal neural relation extraction, where the texts and image posts are crawled from Twitter.
For multimodal NER, we conduct experiments on public  Twitter dataset \textbf{Twitter-2017}~\cite{lu2018visual}, which mainly include multimodal user posts published on Twitter during 2016-2017.

\begin{table}[!tb]
    \caption{
Results of compared models for MRE  in the low-resource setting.
     We report the results of $F1$ score and  adopt $K =1, 5, 10, 20$ (\# examples per class).}

    \label{tab:low_resource_re}
    \footnotesize
    \centering
    \resizebox{0.45\textwidth}{!}{
    \begin{tabular}{c | c | c | c | c | c}
        \toprule

        \textbf{Dataset} & \textbf{Model} 
        & \textbf{$K = 1$}  & \textbf{$K = 5$} & \textbf{$K = 10$} & \textbf{$K = 20$} \\
                \midrule

        \multirow{4}*{MNRE} 
        & BERT ~\cite{devlin2018bert} & 3.67 & 6.27 & 12.65 & 18.94 \\
        & VisualBERT~\cite{li2019visualbert}  & 2.93 & 4.91 & 6.10 & 14.96 \\
        & ViLBERT~\cite{vilbert}  & 3.89 & 7.78 & 13.38 & 18.45 \\
        & MEGA~\cite{multimodal-re} & 9.84 & 11.71 & 15.39 & 20.26 \\
        & {\ours}
        & 12.04 \tiny{\color{red}{(+2.2)}}  
        & 18.54 \tiny{\color{red}{(+6.83)}}
        & 21.09 \tiny{\color{red}{(+5.7)}}
        & 40.93 \tiny{\color{red}{(+20.67)}} \\

        \bottomrule
        
    \end{tabular}}
\end{table}

\subsubsection{Compared Baselines}
We compare our {\ours} with several baseline models for a comprehensive comparison to demonstrate the superiority
of our {\ours}. 
Firstly, we choose the conventional text-based models for comparison to demonstrate the improvement brought by the visual information.
Secondly, we also compare our {\ours} with
\textit{VisualBERT} \cite{li2019visualbert}
and \textit{ViLBERT} \cite{vilbert},  which are pre-trained visual-language model with single-stream structure and two-stream structure respectively.
Besides, we further consider another group of previous SOTA multimodal approaches for multimodal knowledge graph completion models as follows:

\textbf{Multimodal link prediction:}
1) \textit{IKRL} \cite{IKRL}, which extends TransE to learn visual representations of entities and structural information of KG separately;
2) \textit{TransAE} \cite{TransAE}, which combines multimodal autoencoder with TransE to encode the visual and textural knowledge into the unified representation, and the hidden layer of the autoencoder is used as the representation of entities in the TransE model.
3) \textit{RSME} \cite{RSME}, which designs a forget gate with an MRP metric to select valuable images for the multimodal KG embeddings learning.
\textbf{Multimodal RE:}
1) \textit{BERT+SG} is proposed in \cite{multimodal-re} for MRE,  which concatenates the textual representation from BERT with visual features generated by scene graph (SG) tool~\cite{sg}.
2) {\it MEGA}~\cite{multimodal-re} designs the dual graph alignment of the correlation between entities and objects, which is the newest SOTA for MRE.
\textbf{Multimodal NER: }
1) {\it AdapCoAtt-BERT-CRF}~\cite{zhang2018adaptive}, which designs an adaptive co-attention network to induce word-aware visual representations for each word;
2) {\it UMT}~\cite{yu-etal-2020-improving}, which extends Transformer to multi-modal version and incorporates the auxiliary entity span detection module;
3) {\it UMGF}~\cite{zhang-UMGF}, which proposes a unified multimodal graph fusion approach for MNER and achieves the newest SOTA for MNER.

\begin{figure}

    \centering
    \includegraphics[width=0.33\textwidth]{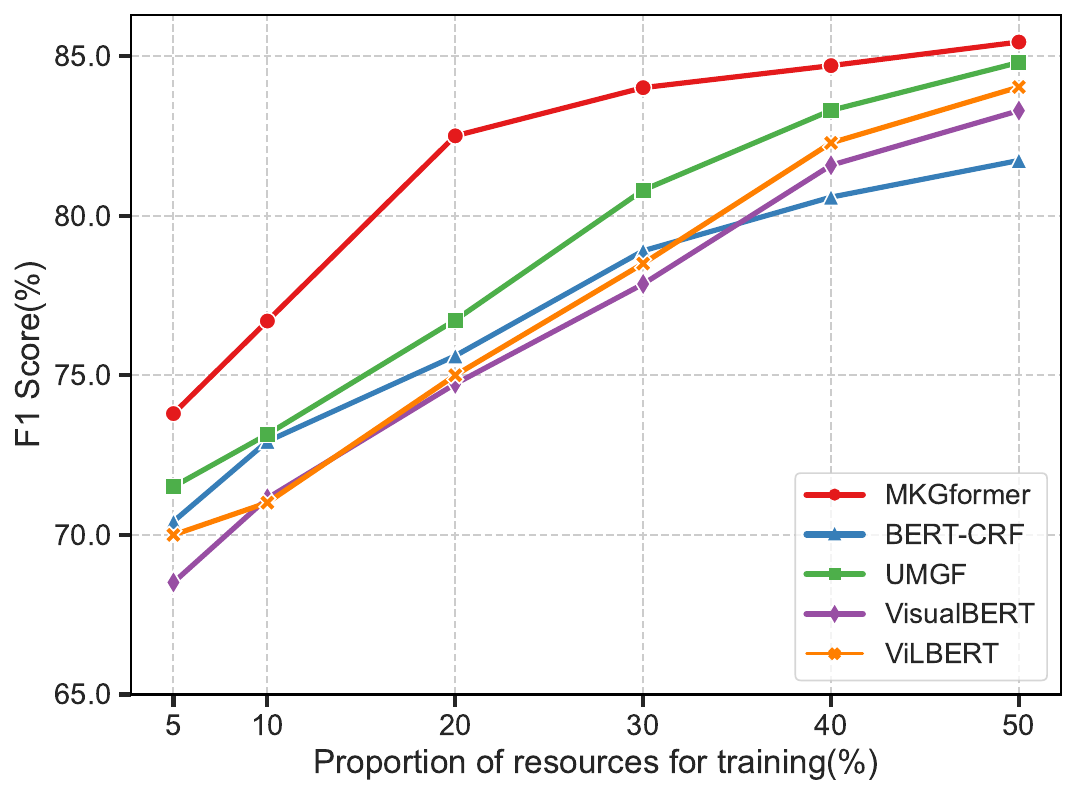}
    \caption{\label{fig:low_resource_ner} Performance of low-resource setting on Twitter-2017 dataset for multimodal NER task.  
    }
\end{figure}

\subsubsection{Settings}
\label{sec:setting}
Notably, we assign the layer of M-Encoder as $L_{M}=3$ and conduct experiments with  BERT\_base and ViT-B/32  \cite{devlin2018bert} for all experiments.
We further conduct extensive experiments in the low-resource setting by running experiments over five randomly sampled $\mathcal{D}_{train}$ for each task and report the average results on test set.
For the few-shot multimodal link prediction and  MRE, we follow the settings of \cite{DBLP:conf/acl/GaoFC20}. 
We adopt  whole images corresponding to entities for multimodal link prediction tasks. 
As for the MRE and MNER tasks, we follow ~\cite{zhang-UMGF} to adopt the visual grounding toolkit~\cite{DBLP:conf/iccv/YangGWHYL19} for extracting local visual objects with top $m$ salience.

\subsection{Overall Performance}
\label{sec:results}

\subsubsection{{Multimodal link prediction}} 
The experimental results in Table~\ref{tab:full_lp} show that incorporating the visual features is generally helpful for link prediction tasks, indicating the superiority of our {\ours}.
Compared to the multimodal SOTA method RSME,
{\ours} has an increase of 1.4\% hits@1 scores and 3.7\% hits@10 scores on FB15k-237-IMG.
We further observe that VisualBERT and ViLBERT perform even worse than SOTA modal RSME.
Note that the implementation of {\ours} is also relatively efficient and straightforward compared with previous knowledge graph embedding approaches \cite{Bordes:TransE} that iteratively query all entities.

\subsubsection{{Multimodal RE and NER}}
From the experimental results shown in Table~\ref{tab:full_re}, we can find that our {\ours} is superior to the newest SOTA models UMGF and MEGA, which improves 1.98\% F1 scores for the Twitter-2017 dataset and  15.44\% F1 scores for the MNRE dataset. 
We further modify the typical pre-trained vision-language model VisualBERT and ViLBERT with [CLS] classifier for the MRE task and CRF classifier for the MNER task to conduct experiments for comparison.
We notice that VisualBERT and ViLBERT perform worse than our methods.
We hold that the poor performance of the pre-trained multimodal model may be attributed to the fact that the pre-training datasets and objects have gaps in information extraction tasks.
This finding also demonstrate that our {\ours} is more beneficial for multimodal MNER and MRE tasks.

\subsection{Low-Resource Evaluation}
Previous experiments illustrate that our methods achieve improvements in standard supervised settings. 
We further report the experimental results in low-resource scenarios compared with several baselines in Figure~\ref{fig:low_resource_ner}, Table~\ref{tab:low_resource_lp} and Table~\ref{tab:low_resource_re}.

\subsubsection{Q1: Does the pre-trained vision-language model successful in the low-resource multimodal KGC? }
As shown in Figure~\ref{fig:low_resource_ner}, Table~\ref{tab:low_resource_lp} and Table~\ref{tab:low_resource_re},
VisualBERT and ViLBERT yield slightly improvements compared with typical unimodal baselines in low-resource settings while obtain worse results than previous multimodal SOTA methods.
It reveals that applying these pre-trained multimodal methods to the multimodal KGC may not always achieve a good performance.
This result may be attributed to the fact that the pre-training dataset and objects of the above visual-language models are less relevant to KGC tasks, which may bias the original language modeling capabilities and knowledge distribution of BERT.

\begin{table}[!tb]
    \caption{Performance when the layer of M-Encoder varies, we take different levels of the activation states into computation, where 1 indicates the bottom layer and 12 indicates the top layer. }
    
    \label{tab:layers}
    \small
    \centering
    \scalebox{0.85}{
    \begin{tabular}{c | c | c | c }
        \toprule

       {\multirow{2}{*}{Methods}}
        &  \multirow{1}{*}{\textbf{FB15k-237-IMG}} 
        &  \multirow{1}{*}{\textbf{MNRE}}
        &  \multirow{1}{*}{\textbf{Twitter-2017}} 
         \\
         \cmidrule{2-4} 
         & \multicolumn{1}{c|}{Hits@10} 
         & \multicolumn{1}{c|}{F1} 
         & \multicolumn{1}{c}{F1} 
         \\
        \midrule

        (layer 12)  & 0.485 & 80.21 & 85.25  \\
        \textbf{(layer 10-12)}  & 0.504 & 81.95  & 87.49  \\
        (layer 7-12) & 0.503 & 82.20 & 87.62  \\
        (All layers)
        & 0.507
        & 82.25
        & 87.60
         \\
        \bottomrule
        
    \end{tabular}}
\end{table}

\subsubsection{Q2: Whether hybrid transformer framework data-efficient?}
Since pre-trained multimodal models do not show promising advantages in low-resource settings, we further analyze whether the hybrid transformer framework is data-efficient.
We hold that leveraging a hybrid transformer framework with similar arithmetic units to fuse entity description and images inside transformers intuitively reduces heterogeneity, playing a more critical role in low-resource settings.
Thus, we compare with previous multimodal KGC SOTA models in low-resource settings.
The experimental results in Figure~\ref{fig:low_resource_ner}, Table~\ref{tab:low_resource_lp} and Table~\ref{tab:low_resource_re} indicate that the performance of {\ours} still outperforms the other baselines, which further proving that our proposed method can more efficiently leverage multimodal data.
This success may be attributed to the prefix-guided fusion module in {\ours} leveraging a form similar to linear interpolation to fuse features in the attention layer, thus, effectively pre-reducing modal heterogeneity.

\subsection{Sensitivity Analysis of M-Encoder Layers}
We assign the M-Encoder with last three  layers of ViT and BERT in the previous experiments.
However, it is intuitive to investigate whether the performance of {\ours} is sensitive to the layers of the M-Encoder.
Thus, we take the M-Encoder in different layers into computation for further analysis.
As shown in Table~\ref{tab:layers}, the performance of {\ours} on FB15k-237-IMG, MNRE and Twitter-2017 only achieve improvements of 0.3\% Hits@10 scores, 0.3\% F1 scores and  0.11\% F1 scores by conducting M-Encoder with all 12 layers. Furthermore, the performance of assigning M-Encoder with only one layer also drops slightly compared with the original result (three layers M-Encoder), showing that M-Encoder applied to higher transformer layers can stimulate knowledge and perform modality fusion for downstream tasks more efficiently.
It also reveals that our approach is not sensitive to the layers of the M-Encoder.

\begin{table*}[ht!]
\caption{The first row shows the split of the relevance of image-text pairs, and the several middle rows indicate representative samples together with their entity-object attention and token-wise similarity in the test set of MNRE datasets, and the bottom five rows in the figure show predicted relation of different approaches on these test samples.
}
\label{tab:re_case}
\resizebox{\linewidth}{!}{
\begin{tabular}[t]{p{9cm}p{9cm}}
\toprule[2pt]
    Relevant Image-text Pair
    & Irrelevant Image-text Pair
\\
\midrule[1pt]

Infinity War Director Confirms \colorbox{red!30}{Hulk} is NOT Afraid of \colorbox{yellow!30}{Thanos}.
&
History beckons as \colorbox{red!30}{Trump} - \colorbox{yellow!30}{Kim} summit kicks off in Singapore.
\\
\raisebox{-0.94\totalheight}{
    \includegraphics[width=0.96\linewidth]{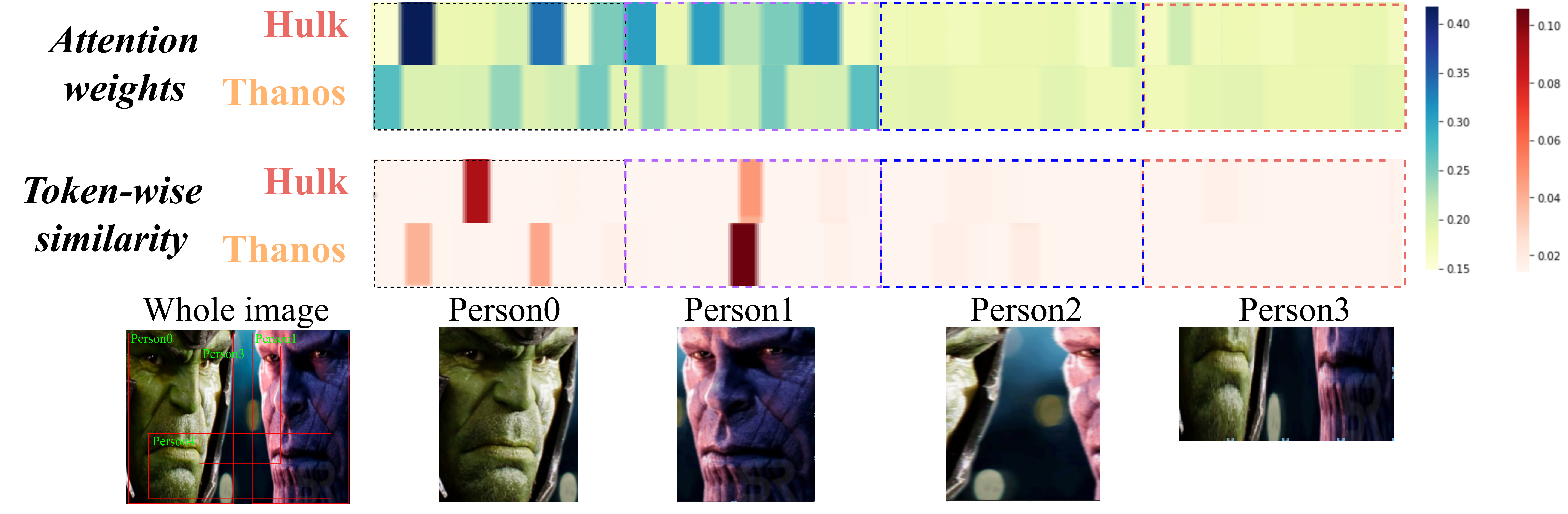} 
} & 
\raisebox{-0.94\totalheight}{
    \includegraphics[width=0.97\linewidth]{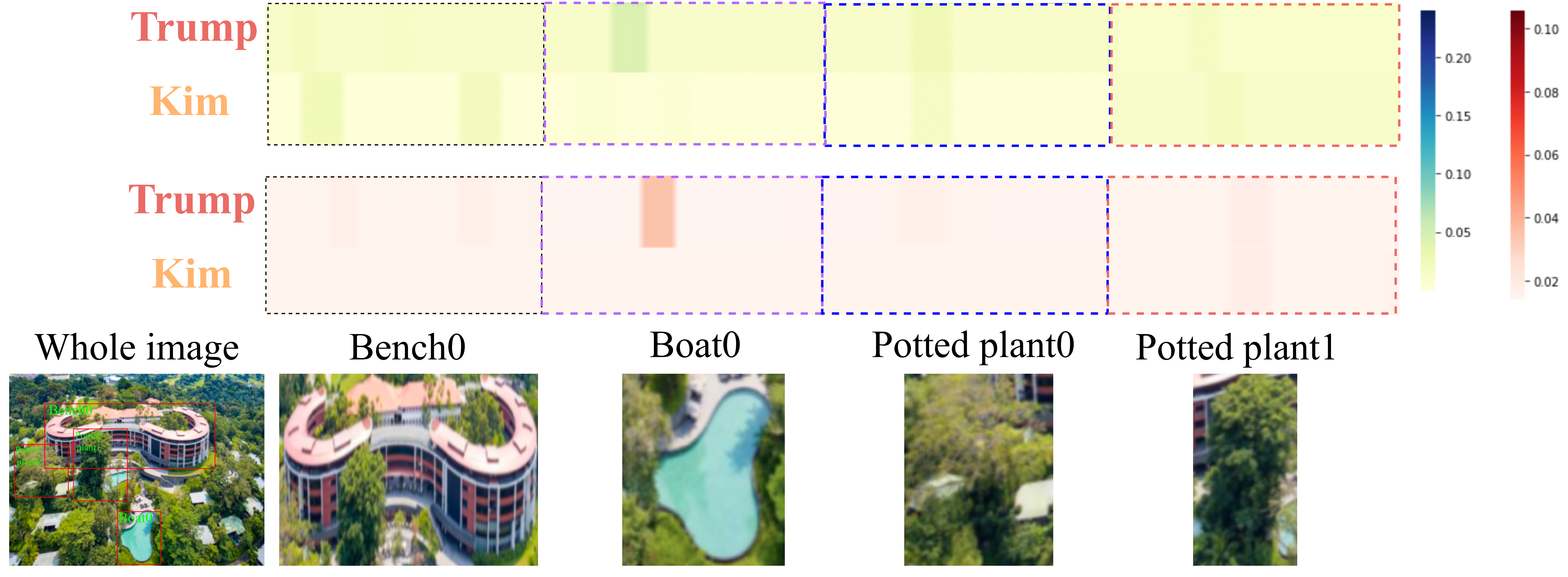} 
}
\\
\textbf{Gold Relations:} 
 per/per/peer & per/per/peer
\\
\midrule[1pt]

\renewcommand\arraystretch{0.7}
\begin{tabular}[t]{lll}
    BERT:  & \colorbox{red!30}{per}/\colorbox{yellow!30}{per}/couple & \xmark\\
    VisualBERT: & \colorbox{red!30}{per}/\colorbox{yellow!30}{per}/peer & \ding{52}\\
    MEGA: & \colorbox{red!30}{per}/\colorbox{yellow!30}{per}/peer & \ding{52}\\
    {Ours}: & \colorbox{red!30}{per}/\colorbox{yellow!30}{per}/peer & \ding{52} \\
    {Ours}(w/o CAF): & \colorbox{red!30}{per}/\colorbox{yellow!30}{per}/peer & \ding{52} \\
\end{tabular} &

\renewcommand\arraystretch{0.7}
\begin{tabular}[t]{lll}
     \colorbox{red!30}{per}/\colorbox{yellow!30}{per}/peer & \ding{52} &\\
    \colorbox{red!30}{per}/\colorbox{yellow!30}{org}/member\_of & \xmark &\\
    \colorbox{red!30}{per}/\colorbox{yellow!30}{loc}/place\_of\_residence & \xmark &\\
     \colorbox{red!30}{per}/\colorbox{yellow!30}{per}/peer & \ding{52} \\
     \colorbox{red!30}{per}/\colorbox{yellow!30}{per}/member\_of & \xmark
\end{tabular} \\

\bottomrule[2pt]
\end{tabular}
}
\end{table*}

\begin{figure}

    \centering
    \includegraphics[width=0.35\textwidth]{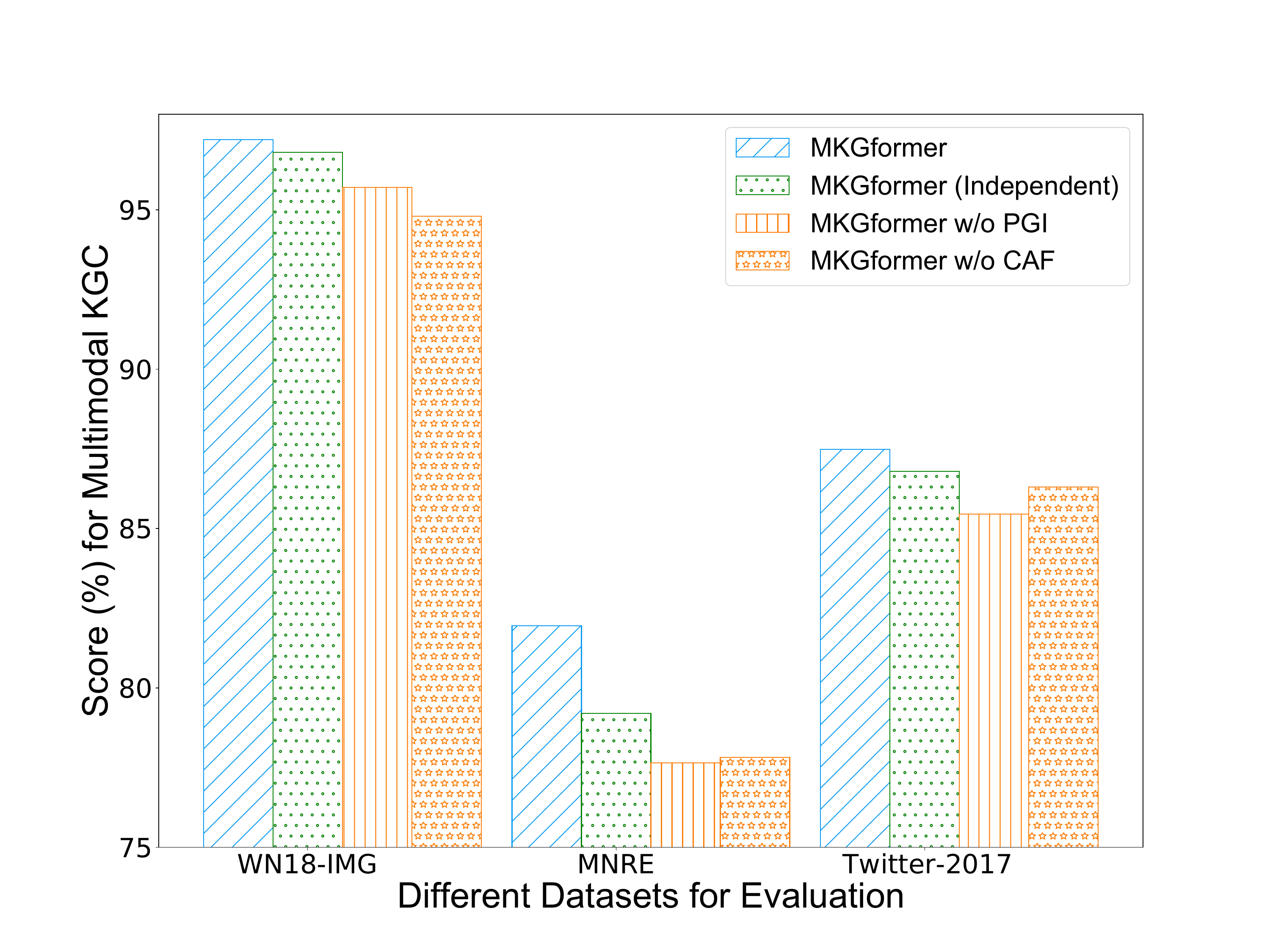}
    \caption{\label{tab:ablation} Ablation study results of MKGformer.}
\end{figure}

\subsection{Ablation Study}
\subsubsection{Ablation Setting}
The ablation study is further conducted to prove the effects of different modules in {\ours}:
\emph{1) (Independent)} indicate that we add extra three layers M-Encoder based on ViT and BERT rather than conduct fusion inside them;
\emph{2) w/o PGI} refers to the model without the  prefix-guided interaction module;
\emph{3) w/o CAF} refers to the model without the whole correlation-aware fusion module.
We report detailed experimental results in  Figure~\ref{tab:ablation} and observe that ablation models both show a performance decay, which demonstrates the effectiveness of each component of our approach.

\subsubsection{Effectiveness of Internal Hybrid Fusion in Transformer}
A specific feature of our method is that we conduct modal fusion inside the dual-stream transformer rather than adding a fusion layer outside the transformer like IKRL~\cite{IKRL}, UMGF~\cite{zhang-UMGF} and MEGA~\cite{multimodal-re}. 
To this end, we add extra three layers M-Encoder based on ViT and BERT to evaluate the impact of the internal fusion mechanism.
We observe that the performance of \emph{{\ours} (Independent)}  drops on three sub-tasks of multimodal KGC, revealing the effectiveness of internal fusion in Transformer.

\subsubsection{Importance of multi-level fusion}
The highlights of our {\ours} is the multi-level fusion in M-Encoder with coarse-grained prefix-guided interaction module and fine-grained correlation-aware fusion module.
We argue that these two parts can mutually reinforce each other: the heterogeneity reduced visual and textual features can help the correlation-aware module better understand fine-grained information. 
On the contrary, the prefix-guided interaction module in the next layer can reduce the modality heterogeneity more gently based on fine-grained fusion in the last layer. 
The results shown in Figure~\ref{tab:ablation} demonstrate that multi-level fusion holds the most crucial role in achieving excellent performance.
At the same time, the case analysis in Table~\ref{tab:re_case} also reveals the impact of the correlation-aware module for alleviating error sensitivity.

\subsection{Case Analysis for Image-text Relevance}
To further analyze the robustness of our method for error sensitivity, we conduct a specific case analysis on the multimodal RE task as indicated in Table~\ref{tab:re_case}.
We notice that  VisualBERT, MEGA, and our method can recognize the relation for the relevant image-text pair. 
Through the  visualization of the case, we can further notice:
1) The attention weights in the prefix-guided interaction module reveal that our model can capture the significant attention between relevant entities and objects. 
2) The similarity matrix also shows that the entity representation from our model is more similar to the corresponding object patch.
Moreover, in the situation that image represents the abstract semantic that is irrelevant to the text, only our method success in prediction due to {\ours} captures the more fine-grained multimodal features.
It is worth noting that another two multimodal baselines fail in irrelevant image-text pairs while text-based BERT and ours still predict correctly.
These observations reveal that irrelevant visual features may hurt the performance, while our model can learn more robust and fine-grained multimodal representation, which is essential for reducing error sensitivity.

\section{Conclusion and Future Work}

In this paper, we present a hybrid  Transformer network for multimodal knowledge graph completion,  which presents M-Encoder with multi-level fusion at the last several layers of ViT and BERT to conduct image-text incorporated entity modeling. 
To the best our knowledge, 
{\ours} is the first work leveraging unified transformer architecture to conduct various multimodal KGC tasks, involving multimodal link prediction, multimodal relation extraction, and multimodal named entity recognition.
Concretely, we propose a prefix-guided interaction module at the self-attention layer to pre-reduce modality heterogeneity and further design a correlation-aware fusion module which realize token-wise fine-grained fusion at the FFN layer to mitigate noise from  irrelevant images/objects.
Extensive experimental results on four datasets demonstrate the effectiveness and robustness of our {\ours}.

In the future, we plan to 
1) apply our approach to more image enhanced natural language processing and information retrieval tasks, such as multimodal event extraction, multimodal sentiment analysis, and  multimodal entity retrieval;
2) apply the reverse version of our approach to boost visual representation with text for CV;
3) extend our approach to pre-training of multimodal KGC.

\begin{acks}
We  want to express gratitude to the anonymous reviewers for their hard work and kind comments. 
This work is funded by NSFC U19B2027/91846204, National Key R\&D Program of China (Funding No.SQ2018YFC000004), Zhejiang Provincial Natural Science Foundation of China (No. LGG22F030011), Ningbo Natural Science Foundation (2021J190), and Yongjiang Talent Introduction Programme (2021A-156-G). Our work is supported by Information Technology Center and State Key Lab of CAD\&CG, ZheJiang University.
\end{acks}


\bibliographystyle{ACM-Reference-Format}
\bibliography{sample-base}


\end{document}